\documentclass{article}
\usepackage{spconf,amsmath,graphicx,hyperref}
\usepackage[ruled,vlined]{algorithm2e}
\usepackage{amsmath}
\usepackage{xcolor} 
\usepackage{booktabs}
\usepackage{multicol} 
\usepackage{float}
\usepackage{amssymb}  

\title{Retrieval-Augmented Self-Taught Reasoning Model with Adaptive Chain-of-Thought for ASR Named Entity Correction}
\name{Junjie An$^{1, 2}$, Jingguang Tian$^{1}$, Tianyi Wang$^{1}$, Yu Gao$^{1, *}$, Xiaofeng Mou$^{1}$, Yi Xu$^{1}$}

\address{$^{1}$ AI Research Institute, Midea Group (Shanghai) Co.,Ltd., China, $^{2}$ Zhejiang University, China}

\vspace{-0.6cm}
\begin{document}
\ninept
\maketitle
\def\thefootnote{*}\footnotetext{Corresponding author.}\def\thefootnote{\arabic{footnote}}
\begin{abstract}
End-to-end automatic speech recognition (ASR) systems frequently misrecognize domain-specific phrases like named entities, which can cause catastrophic failures in downstream tasks. A new family of named entity correction methods based on large language models (LLMs) has recently emerged. However, these approaches have yet to fully exploit the sophisticated reasoning capabilities inherent to LLMs. To bridge this gap, we propose a novel retrieval-augmented generation framework for correcting named entity errors in ASR. Our approach consists of two key components: (1) a rephrasing language model (RLM) for named entity recognition, followed by candidate retrieval using a phonetic-level edit distance; and (2) a novel self-taught reasoning model with adaptive chain-of-thought (A-STAR) that dynamically adjusts the depth of its reasoning based on task difficulty. Experiments on the AISHELL-1 and Homophone datasets demonstrate the effectiveness of our method, which achieves relative reductions in the named entity character error rate of 17.96\% and 34.42\%, respectively, compared to a strong baseline.
\end{abstract}
\vspace{-0.2cm}
\section{Introduction}
\label{sec:intro}
End-to-end automatic speech recognition has achieved significant progress in recent decades, driven by breakthroughs in deep learning~\cite{prabhavalkar2023end}. However, a well-documented vulnerability of these models is their propensity to misrecognize low-frequency, domain-specific terms, such as named entities (NEs) including personal names, locations, and organizations. These entities are often incorrectly transcribed as more common, phonetically similar words.

Numerous post-correction methods have been developed to address the named entity correction (NEC) challenge in ASR. KG-ECO~\cite{cai2023kg} leverages knowledge graphs to enrich both structural and textual representations of entities, thereby enhancing NEC performance. DANCER~\cite{wang2024dancer} mitigates phonetic confusion by employing an efficient masked language model augmented with entity descriptions, which incorporates contextual information to improve recognition accuracy. More recently, research has shifted toward generative error correction (GEC) approaches utilizing LLMs. These methods either prompt LLMs to perform NEC tasks without fine-tuning~\cite{yang2023generative}, or fine-tune LLMs directly using hypothesis–transcription pairs~\cite{ghosh2024failing,pusateri2025retrieval,yamashita25_interspeech}. Recent LLMs, such as OpenAI’s o1~\cite{jaech2024openai} and DeepSeek R1~\cite{guo2025deepseek}, demonstrate that ``slow-thinking" reasoning, which incorporates explicit reasoning stages prior to answer generation, significantly improves task performance. Nevertheless, such structured reasoning paradigms remain largely underexplored in the context of NEC.

It is well-established that training reasoning models requires high-quality rationale data. However, such data are notoriously scarce and costly to obtain, particularly for rare entities and phonetically ambiguous cases. Despite the success of self-training paradigms with self-generated chain-of-thought (CoT) in recent methods such as STAR~\cite{zelikman2024star} and ReST-MCTS~\cite{zhang2024rest}, their application to NEC for ASR remains an open research question.

Retrieval-augmented generation (RAG) has emerged as an effective solution for NEC~\cite{wang2024dancer,ghosh2024failing,pusateri2025retrieval}, leveraging external databases to integrate relevant knowledge. A crucial preliminary step for named entity retrieval is to identify and locate them within ASR transcripts—a process known as named entity recognition (NER). Consequently, the correction capability of an NEC system is fundamentally constrained by the performance of its NER model. However, existing NER models predominantly formulate NER as a sequence tagging task. A critical flaw of character-level tagging is its tendency to overfit mappings between characters and labels during training. Consequently, the resulting tags rely excessively on local character patterns, frequently failing to capture broader sentence-level semantics.

To address these limitations, we propose a retrieval-augmented self-taught reasoning model with adaptive chain-of-thought (RASTAR). we first propose a BERT-based~\cite{devlin2019bert} rephrasing language model (RLM), designed to handle NER by emulating human-like paraphrasing. Rather than relying on memorized local character patterns, RLM reformulates entire sentences through semantic understanding, thereby capturing global semantic information. Building on this, we introduce a novel self-taught reasoning model with adaptive chain-of-thought (A-STAR). It employs self-training techniques to generate CoT rationale training data, bootstrapping its ability to handle challenging NEC cases. Furthermore, it autonomously assesses task complexity and adjusts the depth of reasoning accordingly, significantly reducing inference costs.


\begin{figure*}[htbp]
	\centering
	\includegraphics[width=\textwidth]{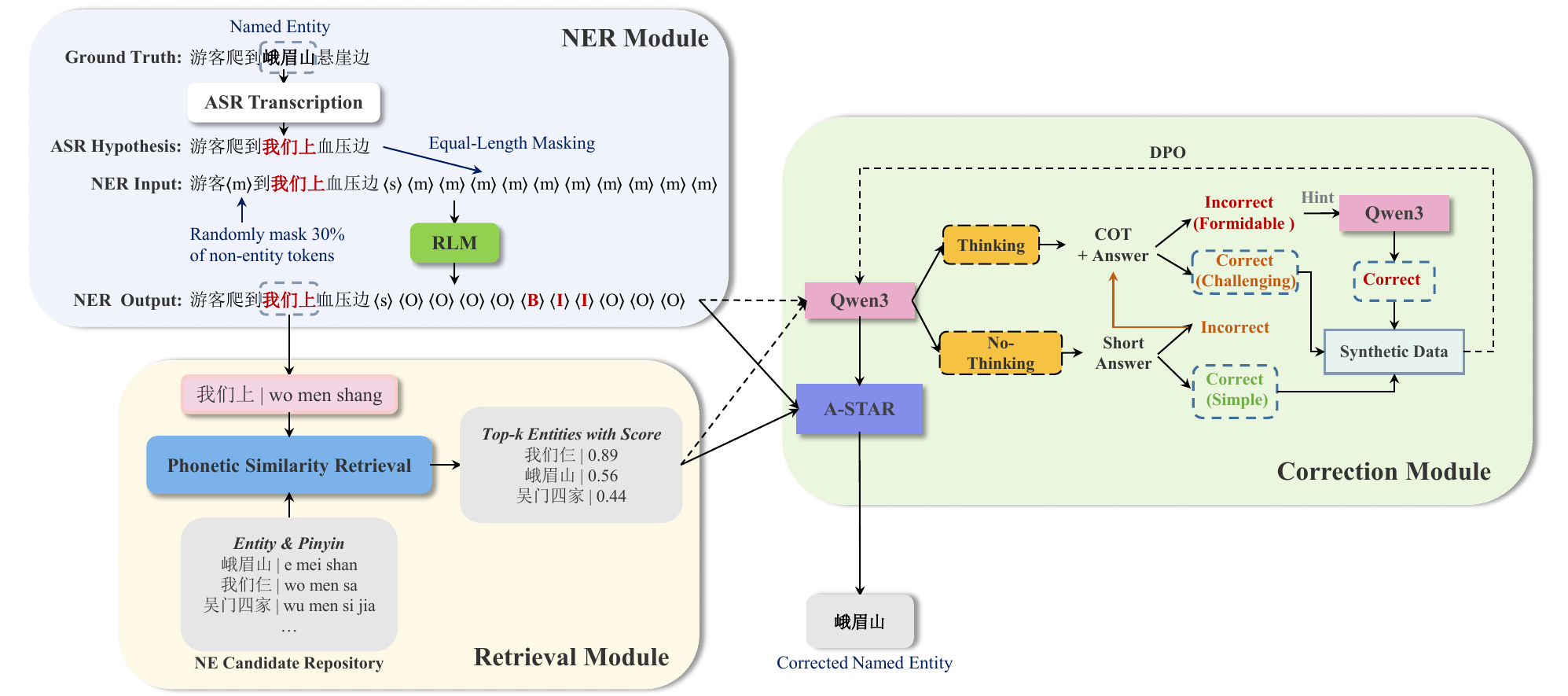}
	\vspace{-15pt}  
	\caption{Architecture of the RASTAR. The system processes ASR transcripts through three sequential modules: (1) \textbf{NER Module} identifies entities $\mathcal{E}^{X}$ from ASR transcripts; (2) \textbf{Retrieval Module} retrieves top-$k$ candidate entities from a NEs candidate repository; (3) \textbf{Correction Module} employs a self-taught reasoning mechanism to select the most appropriate candidate for entity replacement.}
	\label{fig:pipeline}
	\vspace{-10pt}  
\end{figure*}

\vspace{-0.2cm}
\section{Methodology}
\label{sec:Methodology}

\vspace{-0.2cm}
\subsection{Preliminary}

Our RASTAR model addresses the challenges encountered by conventional GEC in handling rare NEs by retrieving relevant NEs from external knowledge bases~\cite{im2025deragec}.
The problem begins with an ASR hypothesis with $n$ tokens represented by $X = \{x_1, x_2, \cdots, x_n\}$, and a pre-defined NEs candidate repository consisting of $C$ entities denoted as $\mathcal{E} = \{\mathbf{e}_1, \cdots, \mathbf{e}_c\}$, where $\mathbf{e}_c$ is an entity composed of a list of character tokens. The goal of RASTAR is to identify the possible set of NEs 
$\mathcal{E}^{X} = (\mathbf{e}_1^{X}, \cdots, \mathbf{e}_M^{X})$ for the given hypothesis $X$, and replace each of them with the correct candidate  selected from $\mathcal{E}$.

RASTAR consists of three core modules. First, the NER module identifies NEs $\mathcal{E}^{X}$ from the given ASR transcripts. Next, the retrieval module fetches the top-k entity candidates for each detected corrupted entity. Finally, the correction module employs A-STAR model to replace the corrupted entity with the most appropriate candidate. The overall architecture is illustrated in Fig.~\ref{fig:pipeline}.

\vspace{-0.2cm}

\subsection{RLM for Named Entity Recognition}
\label{sec:Rephrasing-MLM}
Current state-of-the-art methods formulate NER as a sequence tagging problem and address it by fine-tuning BERT-based models on sentence pairs~\cite{keraghel2024survey}. Given the ASR hypothesis $X$, the model seeks to generate the target sentence $Y = \{y_1, y_2, \cdots, y_n\}$, where $y_i$ is in charge of categorizing character $x_i$ into three classes with the BIO format, following the conditional probability $P(y_i \mid X)$. In this format, \textlangle B\textrangle and \textlangle I\textrangle represent the beginning token and the inside token of an entity, and \textlangle O\textrangle represents non-entity tokens. 

However, the model's character-level mapping paradigm relies heavily on precise input-output alignments, making it vulnerable to speculative behaviors in real-world ASR systems with frequent transcription errors~\cite{liu2024chinese}. For example, as shown in the NER module of Fig.~\ref{fig:pipeline}, the entity \textit{Éméi Shān} (means ``Mount Emei") becomes \textit{wǒ men shàng} (means ``we go up") after ASR transcription---a phonetically similar non-entity unseen during training. This results in recognition failures. The phenomenon can be abstracted as the following training objective degradation:
\begin{equation}
\footnotesize
    P(y_i \mid X) \approx P(y_i \mid x_i)
\end{equation}

Building on the work in~\cite{liu2024chinese}, based on BERT, we propose RLM, a non-auto-regressive phrasing model, which fully leverages the powerful contextual semantic understanding capabilities of pre-trained models.
As shown in the NER module of Fig.~\ref{fig:pipeline}, we concatenate the ASR hypothesis $X$ with a mask sequence of equal length, forming $( x_1, x_2, \cdots, x_n, \left\langle \text{s} \right\rangle, m_1, m_2, \cdots, m_n )$. The model is trained to reconstruct all masked tokens $m_i$. Due to the bidirectional encoding mechanism of BERT, the model seeks to solve the following probability for $y_i,i=1\sim n$:
\begin{equation}
\footnotesize
    P(y_i \mid X) \approx P(y_i \mid X, m_1, m_2, \cdots, m_n).
\end{equation}

To further enhance the model's paraphrasing capability, we employ an auxiliary masked language modeling strategy, which randomly masks a portion of tokens. The model is forced to simultaneously reconstruct the entire sentence while predicting entity labels.

\vspace{-0.2cm}
\subsection{Phonetic-level Retrieval}

Our candidate entities retrieval scheme is inspired by the PED-NEC framework, which employs an efficient phonetic matching mechanism using edit distance~\cite{raghuvanshi2019entity}. The method retrieves candidates from a curated named entity repository $\mathcal{E}$ by computing the conditional probability between the detected entity $\mathbf{e}_m^{X}$ and each candidate $\mathbf{e}_c$:
\begin{equation}
\footnotesize
P(\mathbf{e}_c|\mathbf{e}_m^{X}) = \frac{\phi(\mathrm{Pin}(\mathbf{e}_c),\mathrm{Pin}(\mathbf{e}_m^{X}))}{\sum_{{\mathbf{e}^{\prime}_c}\in\mathcal{E}}\phi(\mathrm{Pin}(\mathbf{e}^{\prime}_c),\mathrm{Pin}(\mathbf{e}_m^{X}))}
\end{equation}
where $\mathrm{Pin}(\cdot)$ denotes the romanization for standard Mandarin Chinese and $\phi$ measures the phonetic similarity between $\mathbf{e}_c$ and $\mathbf{e}_m^{X}$ based on the normalized edit distance.

For each detected span $\mathbf{e}_m^{X}$, we rank the candidates in $\mathcal{E}$ by their conditional probability and selects the top-k candidates: $\mathcal{E}_{\mathbf{e}_m^{X}}=\{\tilde{\mathbf{e}}_1, \cdots, \tilde{\mathbf{e}}_k\}$, which will serve as a context reference for the correction module.

\vspace{-0.2cm}
\subsection{A-STAR}

We form the input to the reasoning model by concatenating the ASR hypothesis $X$, the top-k entity candidates $\mathcal{E}_{\mathbf{e}_m^{X}}$, and a task-specific prompt:
$Q=\{X,\mathcal{E}_{\mathbf{e}_m^{X}},\mathbf{prompt}\}$.

Current reasoning models support switching between thinking and nothinking modes~\cite{chen2025towards}. For example, Qwen3~\cite{yang2025qwen3} allows mode transition via prompt flags (\texttt{/think}) or an API parameter (\texttt{enable\_thinking}).
However, these methods require explicit mode selection, lacking autonomous judgment about when slow-thinking is needed~\cite{zhang2025othink, fang2025thinkless, zhang2025adaptthink}. This often results in unnecessarily long chains of thought even for simple NEC tasks, producing redundant text and substantial computational overhead.

Simultaneously, enabling models to effectively master the reasoning paradigms specific to NEC tasks necessitates large-scale CoT data, which remains scarce and costly to annotate manually. A promising alternative involves leveraging models to autonomously generate CoT rationales from non-CoT data through self-training~\cite{zelikman2024star}, an approach that offers sustainable self-improvement and extensibility.


To address these challenges, we propose A-STAR, which features self-distillation without the need for manually annotated COT data, and is capable of autonomously switching between thinking modes according to problem complexity. As illustrated in the correction module of Fig.~\ref{fig:pipeline}, we begin by processing the problem dataset $Q$ through the reference model to generate both COT responses ($A_{\text{think}}$) and direct responses with brief explanations ($A_{\text{nothink}}$).
Subsequently, we classify problem difficulty by comparing $A_{\text{nothink}}$ responses with ground truth labels. Problems correctly answered in nothinking mode are categorized as ``Simple", forming the subset $Q_{\mathrm{S}}$ with corresponding responses $A_{\mathrm{S}}$.

For the remaining problems in $Q$, we evaluate their corresponding responses in $A_{\text{think}}$ and perform secondary classification based on answer correctness: problems solved correctly are categorized as ``Challenging" ($Q_{\mathrm{C}}$), while those remaining unsolved even in thinking mode are labeled as ``Formidable" ($Q_{\mathrm{F}}$). Their corresponding responses form subsets $A_{\mathrm{C}}$ and $A_{\mathrm{F}}$, respectively.
Inspired by the works of STAR~\cite{zelikman2024star}, we address $Q_{\mathrm{F}}$ by incorporating corrective hints containing ground-truth answers into the prompts. This allows the model to regenerate refined responses through reflection, followed by rejection sampling to produce the final response set $A'_{\mathrm{F}}$.

Finally, we synthesize preference pairs from $A_{\mathrm{S}}$, $A_{\mathrm{C}}$, and $A'_{\mathrm{F}}$ to form synthetic data for self-improvement. Specifically, we treat responses generated in nothinking mode as preferred responses for simple problems ($Q_S$), while the dispreferred responses pairs are those produced in thinking mode. For harder problems ( $Q_C$ and $Q_F$), this preference relationship is reversed. We employ Direct Preference Optimization (DPO) ~\cite{rafailov2023direct} for self-training, with the training loss expressed by:
\begin{equation}
\footnotesize
\mathcal{L}_{\text{DPO}}(\pi_\theta; \pi) = 
-\mathbb{E}_{(q,y^+,y^-) \sim \mathcal{D}} 
\left[ 
\log \sigma\left( 
\beta \left( r_\theta(q, y^+) - r_\theta(q, y^-) \right)
\right) 
\right]
\end{equation}
where $\pi_\theta$ denotes the policy model to be optimized, $\pi$ denotes the reference model, $q \in Q$ denotes the input, $\mathcal{D}$ denotes the preference dataset containing triples $(q, y^+, y^-)$, $y^+$ represents the preferred response, and $y^-$ represents the dispreferred response, $r_\theta(q, y) = \log \frac{\pi_\theta(y \mid q)}{\pi(y \mid q)}$ denotes the implicit reward function, 
$\beta$ denotes the temperature parameter controlling deviation from the reference $\pi$. The complete workflow of A-STAR is detailed in algorithm~\ref{algorithm 1}.

\begin{algorithm}[h]
\label{algorithm 1}
\caption{A-STAR}
\KwIn{Policy model $\pi_\theta$, problem dataset $Q$, labels $Y$}
\BlankLine
1.\ $A_{\text{nothink}}  \gets {\pi_{\theta-{\text{nothink}}}}(Q)$\;
2.\ $Q_S\gets \{q\!\in\!Q : A_{\text{nothink}}(q)=Y_q\},Q_R  \gets Q \setminus Q_S$\;
3.\ $A_S\gets \{A_{\text{nothink}}(q)\!\in\!A_{\text{nothink}} : q\!\in\!Q_S\}$\;
4.\ $A_{\text{think}} \gets {\pi_{\theta-{\text{think}}}}(Q_R)$\;
5.\ $Q_C\gets\{q\!\in\!Q_R : A_{\text{think}}(q)=Y_q\},\;
      Q_F\gets Q_R \setminus Q_C$\;
6.\ $A_C\gets \{A_{\text{think}}(q)\!\in\!A_{\text{think}} : q\!\in\!Q_C\}$\;
7.\ $A'_F \gets {\pi_{\theta-{\text{think}}}}(Q_F,\mathrm{hint},Y)$ \quad//rejection sampling
\BlankLine
8.\ Build DPO pairs: \\
    \quad $\mathcal{D} \xleftarrow{\text{sampled}}
         \{(q, A'_{F}(q), A_{\text{nothink}}(q)) : q\in Q_F\}\ \cup$\\
    \quad\quad\quad\quad\quad
         $\{(q, A_C(q), A_{\text{nothink}}(q)) : q\in Q_C\}\ \cup$\\
    \quad\quad\quad\quad\quad
         $\{(q, A_{\text{S}}(q), A_{\text{think}}(q)) : q\in Q_S\}$
\BlankLine
9.\ $\pi'_\theta \gets \text{DPO-Train}(\pi_\theta, \mathcal{D})$\;
\Return $\pi'_\theta$\;
\end{algorithm}


\begin{table}[htbp]
  \centering
  \vspace{-20pt}
  \caption{NER Performance on the AISHELL-1 and Homophone Test Sets. DANCER refers specifically to its NER component.}
  \ninept                       
  \setlength\tabcolsep{3pt}
  \begin{tabular}{lcccccc}
    \toprule
    \textbf{Model} & \multicolumn{3}{c}{\textbf{AISHELL-1 (\%)}} &
                   \multicolumn{3}{c}{\textbf{Homophone (\%)}} \\
    \cmidrule(lr){2-4} \cmidrule(lr){5-7}
     & \textbf{recall} & \textbf{precision} & \textbf{F1}
     & \textbf{recall} & \textbf{precision} & \textbf{F1} \\
    \midrule
    DANCER & 92.57 & 91.39 & 91.97 & 71.19 & 78.84 & 74.82 \\
    RLM   & \textbf{95.55} & \textbf{92.84} & \textbf{94.18}
           & \textbf{83.33} & \textbf{84.49} & \textbf{83.91} \\
    \bottomrule
  \end{tabular}
  \label{tab:NER_performance}
  \vspace{-10pt}  
\end{table}

\vspace{-0.2cm}
\section{EXPERIMENT}

\begin{table*}[htbp]
    \centering
    \caption{NEC Performance on the AISHELL-1 and Homophone Test Sets.}
    \begin{tabular}{lcccccccc}
        \toprule
        \textbf{Model} & \multicolumn{4}{c}{\textbf{AISHELL-1 Test Set (\%)}} & \multicolumn{4}{c}{\textbf{Homophone Test Set (\%)}} \\
        \cmidrule(lr){2-5} \cmidrule(lr){6-9}
         & \textbf{CER} & \textbf{NNE-CER} & \textbf{NE-CER} & \textbf{NE Recall} & \textbf{CER} & \textbf{NNE-CER} & \textbf{NE-CER} & \textbf{NE Recall} \\
        \midrule
        DANCER & 4.29 & 4.00 & 7.57 & 85.84 & 7.17 & 5.35 & 11.33 & 79.84 \\
        DANCER (RLM) & \textbf{4.22} & 4.01 & \textbf{6.81} & \textbf{87.12} & \textbf{7.01} & \textbf{5.27} & \textbf{10.97} & 79.83 \\
        \midrule
        Qwen3 (8B)-nothinking & 4.27 & \textbf{4.04} & 7.14 & 86.94 & 6.79 & 5.43 & 9.91 & 78.91 \\
        Qwen3 (8B)-thinking & 4.24 & 4.07 & 6.47 & 89.22 & 6.15 & 5.27 & 8.14 & 84.57 \\
        Qwen3 (8B)-SFT & \textbf{4.18} & 4.05 & \textbf{6.01} & 89.15 & 6.31 & 5.27 & 8.67 & 82.54 \\
        RASTAR (8B) & 4.21 & 4.05 & 6.21 & \textbf{89.33} & \textbf{6.04} & 5.43 & \textbf{7.43} & \textbf{85.80} \\
        \midrule
        Qwen3 (0.6B)-nothinking & 4.69 & 4.08 & 11.52 & 78.97 & 8.25 & 5.89 & 14.69 & 73.39 \\
        Qwen3 (0.6B)-thinking & 4.27 & \textbf{4.04} & 7.14 & 86.94 & 7.39 & 5.43 & 12.04 & 74.57 \\
        Qwen3 (0.6B)-SFT & 4.28 & 4.06 & 7.07 & 87.70 & 7.12 & 5.43 & 10.97 & 76.96 \\
        RASTAR (0.6B) & \textbf{4.27} & 4.06 & \textbf{6.93} & \textbf{87.92} & \textbf{6.63} & \textbf{5.43} & \textbf{9.38} & \textbf{79.42} \\
        \bottomrule
    \end{tabular}
    \label{tab:NEC_performance}
    \vspace{-9pt}
\end{table*}

\vspace{-0.2cm}
\subsection{Dataset, Evaluation Metrics and Baseline}

Our experimental setup is consistent with the DANCER~\cite{wang2024dancer}. Evaluations were performed on the AISHELL-1 benchmark~\cite{bu2017aishell}, along with a specialized Homophone test set comprising 115 utterances with high phonetic confusion, sampled from AISHELL-1. We additionally utilized the AISHELL-NER data set~\cite{chen2022aishell}, which was constructed based on AISHELL-1, to obtain NER annotation information. The named entity candidate repository $\mathcal{E}$ contained 16,168 distinct entities compiled from the complete training, development, and test sets of AISHELL-1.

We evaluate the NER model on three metrics: recall, precision, and F1.
For NEC methods, we employ four metrics: 1) CER (Character Error Rate), measuring overall character-level accuracy; 2) NNE-CER (Non-Named Entity CER), evaluating error rate in non-entity segments; 3) NE-CER (Named Entity CER), assessing error rate within entity segments; and 4) NE-Recall, quantifying the recall of correctly recognized entities.

To ensure a fair comparison, we employ the same Conformer-based ASR model (trained on AISHELL-1) used in DANCER~\cite{wang2024dancer} to generate hypotheses. NEC experiments are then conducted on these ASR outputs. We reproduced the results of DANCER and established it as a strong baseline. 

\vspace{-0.2cm}
\subsection{Implementation Details}

We employ the bert-base-chinese pre-trained model~\cite{devlin2019bert} for the NER module.
To enhance the model's entity recognition capability in practical scenarios (with ASR transcription error as noise), we utilize n-best ASR transcriptions of the utterances from the AISHELL-1 training set as additional noisy data to train our RLM model. Here we set n=10.
To assign BIO tags to the n-best hypotheses, we align them with the ground truth labels from the AISHELL-NER corpus.
Subsequently, we concatenate the ASR hypotheses and their corresponding equal-length BIO tags in the manner described in Section~\ref{sec:Rephrasing-MLM} to form our NER training set. We then apply complete masking to all BIO tags segment while implementing partial masking on the ASR hypothesis segment.
Through empirical study of masking strategies and mask rates, we ultimately randomly mask 30\% of the non-entity tokens in the ASR hypotheses segment, as shown in the NER module of Fig.~\ref{fig:pipeline}. The loss function for our RLM model is cross-entropy loss, computed only on the masked tokens.

For the phonetic-level retrieval module, through comparative studies we ultimately set $k=3$, as an excessively large $k$ would introduce irrelevant or weakly related entities, interfering with the model's correction performance~\cite{im2025deragec}.

We employ the Qwen3 series models (0.6B and 8B variants) for the A-STAR module in our experiments~\cite{yang2025qwen3}. The DPO datasets for fine-tuning consisted of 1,800 high-quality samples (from Algorithm~\ref{algorithm 1}) for Qwen3-8B and 6,880 samples for Qwen3-0.6B.
In both datasets, the data comprises an equal mix (50\% each) of preferred and dispreferred samples from no-thinking responses. We utilize the \texttt{llamafactory} toolkit for model fine-tuning~\cite{zheng2024llamafactory}, with training conducted on 8 $\times$ 40GB NVIDIA H20 GPUs. We set the temperature parameter $\beta=0.1$. The 8B variant is trained for 10 epochs while the 0.6B variant undergoes 2 training epochs, using a learning rate of $5\mathrm{e}^{-6}$ and gradient accumulation steps set to 8.

\vspace{-0.2cm}
\subsection{Results}
\vspace{-0.2cm}
\subsubsection{Main Results}

Experimental results for the NER module and the overall correction performance are presented in Table~\ref{tab:NER_performance} and Table~\ref{tab:NEC_performance}, respectively. As shown in Table~\ref{tab:NER_performance}, RLM outperforms the tagging-based NER model from DANCER, achieving absolute F1 score improvements of 2.21\% and 9.09\% on the AISHELL-1 and homophone test sets, respectively. Furthermore, replacing DANCER's NER module with RLM—as indicated in the first two rows of Table~\ref{tab:NEC_performance}—yields relative NE-CER reductions of 10.03\% and 3.18\% on the AISHELL-1 and Homophone test sets, respectively. These results demonstrate that our RLM model more effectively leverages contextual semantic information, even in the presence of ASR errors.

Table~\ref{tab:NEC_performance} presents a comparative analysis of LLM-based NEC performance across two parameter scales (0.6B and 8B). The results show that reasoning (``thinking" mode) consistently improved NEC performance over the ``nothinking" mode in the pre-trained Qwen3 model. Following the approach in ~\cite{ghosh2024failing,pusateri2025retrieval,yamashita25_interspeech}, we perform supervised fine-tuning (SFT) of the Qwen3 model on the AISHELL-1 hypothesis–transcription pair dataset. We observe that the fine-tuned Qwen3 loses its reasoning capability, the knowledge learned from the in-domain dataset leads to limited improvements compared to the pre-trained model. Finally, our proposed method demonstrates clear performance advantages, particularly on the challenging Homophone test set, where it achieves relative reductions of 17.96\% and 34.42\% in NE-CER compared to the DANCER.



\vspace{-0.3cm}
\subsubsection{Reasoning Efficiency Analysis}

Table~\ref{tab:cot} summarizes the average CoT length and the ratio of nothinking mode usage for both the Qwen3 and the RASTAR models (0.6B/8B variants). The CoT length for RASTAR was calculated as the average of the outputs from both thinking and nothinking modes. The results show that RASTAR (8B) reduces tokens by 30\% and 21\% on the AISHELL-1 and Homophone test sets, respectively. The ratio of nothinking mode usage shows a consistent trend, and is more evident (40\%) on the relatively simpler AISHELL-1 dataset. This demonstrates that our model can autonomously assess task complexity and switch between thinking modes accordingly, significantly reducing reasoning costs.

However, the same methodology shows constrained compression efficiency for the 0.6B model, with only 3\% of tokens reduction on the Homophone test set.
This indicates that smaller-parameter models struggle to maintain comparable accuracy when reasoning chains are compressed.
Table~\ref{tab:NEC_performance} provides additional validation: on AISHELL-1 test set, the 0.6B model exhibits a 38\% relative decrease in NE-CER when utilizing thinking mode, compared to merely a 9\% reduction for the 8B model. This disparity underscores the greater dependency of smaller models on elaborated reasoning processes for performance gains~\cite{ho2022large}.
Above results collectively indicate that larger models can effectively solve most problems through abbreviated reasoning chains, demonstrating superior compressibility. In contrast, smaller models still require extended intermediate reasoning steps to obtain substantial performance improvements, resulting in stronger reliance on ``slow-thinking" methodologies—which fundamentally constrains their compression potential.

\begin{table}[H]
    \centering
    \vspace{-10pt}
    \caption{Statistics of Mode-Averaged CoT Length and Nothinking Ratio on the AISHELL-1 and Homophone Test Sets.}
    \setlength\tabcolsep{3pt}
    \begin{tabular}{lcccc}
        \toprule
        \textbf{Model} & \multicolumn{2}{c}{\textbf{AISHELL-1}} & \multicolumn{2}{c}{\textbf{Homophone}} \\
        \cmidrule(lr){2-3} \cmidrule(lr){4-5}
         & \textbf{Tokens} & \textbf{Ratio} & \textbf{Tokens} & \textbf{Ratio} \\
        \midrule
        Qwen3 (8B)-thinking  & $405$ & $0$ & $449$ & $0$ \\
        RASTAR (8B) & $284(\downarrow 30\%)$ & 40\% & $355 (\downarrow 21\%)$ & 24\% \\
        \midrule
        Qwen3 (0.6B)-thinking  & $418$ & 0 & $389$ & 0 \\
        RASTAR (0.6B) & $375(\downarrow 10\%)$ & 6\% & $378(\downarrow 3\%)$ & 2\% \\
        \bottomrule
    \end{tabular}
    \label{tab:cot}
    \vspace{-10pt}
\end{table}

\vspace{-0.3cm}
\section{CONCLUSION}

In this paper, we propose a comprehensive framework for NEC in end-to-end ASR systems, addressing key limitations in both NER and GEC. We introduce the RLM, which enhances the robustness of NER under noisy ASR transcriptions by leveraging contextual semantic understanding beyond strict token alignment. Furthermore, we present the A-STAR, which integrates phonetic retrieval with a “slow-thinking” reasoning mechanism to effectively handle rare and challenging named entities.
A-STAR enables two key capabilities: self-improvement through training on self-distilled data, and adaptively switches thinking modes based on task complexity which significantly reduces reasoning costs. Experimental results demonstrate that our method achieves substantial improvements in both NER and NEC performance, particularly in phonetically challenging cases, while also compressing reasoning chains for larger models without sacrificing accuracy.

\vfill
\pagebreak

\bibliographystyle{IEEEbib}
\bibliography{strings,refs}

\end{document}